\documentclass[11pt]{article}
     \usepackage[preprint]{neurips_2020}

\usepackage{natbib}
\setcitestyle{numbers,square}
\usepackage[utf8]{inputenc} 
\usepackage[T1]{fontenc}    
\usepackage{hyperref}       
\usepackage{url}            
\usepackage{booktabs}       
\usepackage{amsfonts}       
\usepackage{nicefrac}       
\usepackage{microtype}      

\usepackage{graphicx}    
\usepackage{amsmath}
\usepackage{amsfonts}
\usepackage{amssymb}
\usepackage{bbm}
\usepackage{algorithm}
\usepackage{algorithmic}

\title{Speed up the inference of diffusion models via shortcut MCMC sampling}

\author{%
  Gang Chen
    \\
  Department of Computer Science\\
  SUNY at Buffalo\\
  Buffalo, NY 14260 \\
  \texttt{newhorizontal@gmail.com} \\
}

\begin{document}         
\date{}

\maketitle
\begin{abstract}
Diffusion probabilistic models have generated high quality image synthesis recently. However, one pain point is the notorious inference to gradually obtain clear images with thousands of steps, which is time consuming compared to other generative models. In this paper, we present a shortcut MCMC sampling algorithm, which balances training and inference, while keeping the generated data's quality. In particular, we add the global fidelity constraint with shortcut MCMC sampling to combat the local fitting from diffusion models. We do some initial experiments and show very promising results. Our implementation is available at https://github.com//vividitytech/diffusion-mcmc.
\end{abstract}

\section{Introduction}
Leveraging deep generative models to generate high quality images has becoming the dominant approach in machine learning community. 
For example, generative adversarial networks (GANs) \cite{Goodfellow2014}, PixelCNN \cite{Oord2016} and variational autoencoders \cite{Kingma2014}
have shown impressive image and speech synthesis results. Diffusion probabilistic models \cite{DicksteinW15} have recently gained popularity over a variety of applications on computer vision and machine learning domain. And it also obtains state-of-the-art Inception score and FID score \cite{Jonathan20,Nichol21,YSong21} on image generation, as well as best results on density estimation benchmarks \cite{Kingma21}. Diffusion models are well defined with Markov chain assumption and are efficient to train. But it is time consuming to generate high quality images, which may take thousands of steps to the best of our knowledge. This paper presents an approach to speed up the inference of diffusion models. Instead of thousands of steps to produce samples, we constrain the number of inference steps, which can be randomly sampled from these thousand steps (we call shortcut MCMC) and then generate images to match the data. Both denoising diffusion probabilistic models (DDPMs) and variational diffusion models (VDMs) train a similar denoising deep nets, which focus on local model characteristics and thus long sampling steps needed to produce high quality images. 

Compared to VDMs, we introduce the shortcut MCMC sampling and add the fidelity term in the loss function so that the final synthesized image match the original data. This new fidelity term is more like a global constraint and quality control while generating images in a shortcut manner. Thus, our method can balance the training and inference stages, and mitigates the inference burden significantly. We do some initial analysis and show promising results on synthesis dataset.

\section{Background}
The diffusion models \cite{DicksteinW15,Jonathan20} are composed of forward process and reverse (backward) process. Given the data $x_0 \sim q(x_0)$, the forward (diffusion) process follows a Markov chain
\begin{align}\label{eq:fw}
q(\mathbf{x}_t | \mathbf{x}_{0}) = \mathcal{N} (\mathbf{x}_{t}, \alpha_t \mathbf{x}_{0} + \sigma_t \mathbf{I}), \quad q(\mathbf{x}_{1:T} | \mathbf{x}_0) = \prod_{t=1}^T q( \mathbf{x}_t |\mathbf{ x}_{t-1})
\end{align}
where $\alpha_t = \sqrt{1- \sigma_t^2}$, and $(\alpha_t, \sigma_t)$ is the signal and noise pair at time step $t$. the Markov chain $q( \mathbf{x}_t |\mathbf{ x}_{t-1})$ is Gaussian
\begin{align}
q( \mathbf{x}_t |\mathbf{ x}_{t-1}) = \mathcal{N}(\alpha_{t|t-1}, \sigma^2_{t|t-1} \mathbf{I})
\end{align}
where $\alpha_{t|t-1} = \alpha_{t}/\alpha_{t-1}$ and $\sigma^2_{t|t-1} = \sigma_t^2 - \alpha^2_{t|t-1} \sigma^2_{t-1}$ according to VDMs \cite{Kingma21}.
The reverse (or backward) process is to learn $p(\mathbf{x}_0) = \int p(\mathbf{x}_{0:T}) d \mathbf{x}_{1:T})$, where $p(\mathbf{x}_T)$ is Gaussian $\mathcal{N}(\mathbf{x}_T; 0, \mathbf{I} )$:
\begin{align}\label{eq:bw}
p(\mathbf{x}_{t-1} | \mathbf{x}_{t}) = \mathcal{N} (\mathbf{x}_{t-1}; \mu_{\theta}(x_t, t), \sigma_\theta(x_t, t )), \quad p(\mathbf{x}_{0:T}) = p(x_T) \prod_{t=1}^T p( \mathbf{x}_{t-1} |\mathbf{ x}_{t})
\end{align}

\begin{figure*}[t!]
\begin{tabular}{c}
\includegraphics[trim=440mm 2.0mm 192mm 3mm, clip, width=14.2cm]{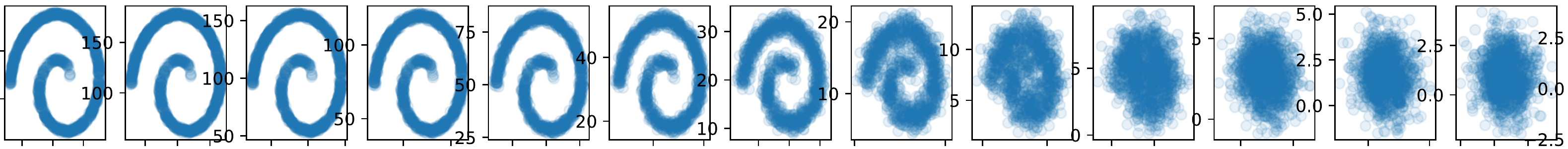}
\end{tabular}
\caption{The noised data with increasing noise level until random Gaussian distribution.}
\label{Fig:noise}
\end{figure*}
Fig \ref{Fig:noise} shows the examples while increasing noise signal over the original data. By optimizing the variational lower bound, VDMs \cite{Kingma21} chooses the conditional model distributions below
\begin{align}\label{eq:kl}
p(\mathbf{x}_{t-1}| \mathbf{x}_t ) = q(\mathbf{x}_{t-1} | \mathbf{x}_t, \mathbf{x}_0)
\end{align}
which can be induced according to the KL divergence. In the inference stage, we can replace $\mathbf{x}_0$ with its prediction $\hat{\mathbf{x}_0}(x_t; t)$ using denoising diffusion models.

\section{Model}
In this section, we will introduce our approach based on the variational lower bound and the shortcut MCMC sampling to skip multiple steps to speed up inference. We consider the finite time steps and it can be easily extended to continuous scenario. 
\subsection{Objective lower bound}
In the case of finite $T$ steps, we maximize the variational lower bound of marginal likelihood below
\begin{align}
\mathcal{L}(\mathbf{x}_0; \theta) =  E_{q(z| \mathbf{x} )} [ \log p(\mathbf{x}_0 |z)] - D_{KL} (q(\mathbf{x}_T | \mathbf{x}_0) ||p(\mathbf{x}_T)) - \sum_{t=2}^T D_{KL} ( q(\mathbf{x}_{t-1} | \mathbf{x}_{t}, \mathbf{x}_0) || \log p(\mathbf{x}_{t-1} | \mathbf{x}_t) )
\end{align}
where $z=(x_1,x_2,...,x_T)$, and for detail induction, please refer Appendix A.
Compared to VDMs, we have an additional fidelity term $\mathbb{E}_{q} \log p(\mathbf{x}_0 | z)$, which maps the latent (prior) Gaussian noise to data distribution. This is similar to GANs model, which can generate data from latent distribution. However, for diffusion model, it depends on the hyperparameter $T$ that will take thousands of steps (e.g. $T =1000$) to produce synthesized data. In other words, it is 3 orders of magnitude slower than GANs when both use the similar deep neural nets architecture in the inference stage.
 
As for the diffusion loss, it leverages KL-divergence to match $p(\mathbf{x}_{t-1} | \mathbf{x}_t)$ with the forward process posterior $q(\mathbf{x}_{t-1} | \mathbf{x}_{t}, x_0)$. Since both the forward posterior and $p(\mathbf{x}_{t-1} | \mathbf{x}_t )$ are Gaussians, with same variance assumption, then the KL loss can be minimized using the deep denoise model
\begin{align}
D_{KL} ( q(\mathbf{x}_{s} | \mathbf{x}_{t}, \mathbf{x}_0) || \log p(\mathbf{x}_{s} | \mathbf{x}_t) ) = \frac{1}{2}( \frac{\alpha_s^2}{\sigma_s^2}  - \frac{\alpha_t^2}{\sigma_t^2} ) || \epsilon - \hat{\epsilon}_\theta( \mathbf{x}_t, t ) ||^2
\end{align}
where $0<s < t \le T$, and $(\alpha_s, \sigma_s)$ and $(\alpha_t, \sigma_t)$ are signal and noise pairs respectively at time step $s$ and $t$.

\begin{figure*}[t!]
\begin{tabular}{c}
\includegraphics[trim=35mm 60.0mm 35mm 30mm, clip, height=6.5cm, width=14.2cm]{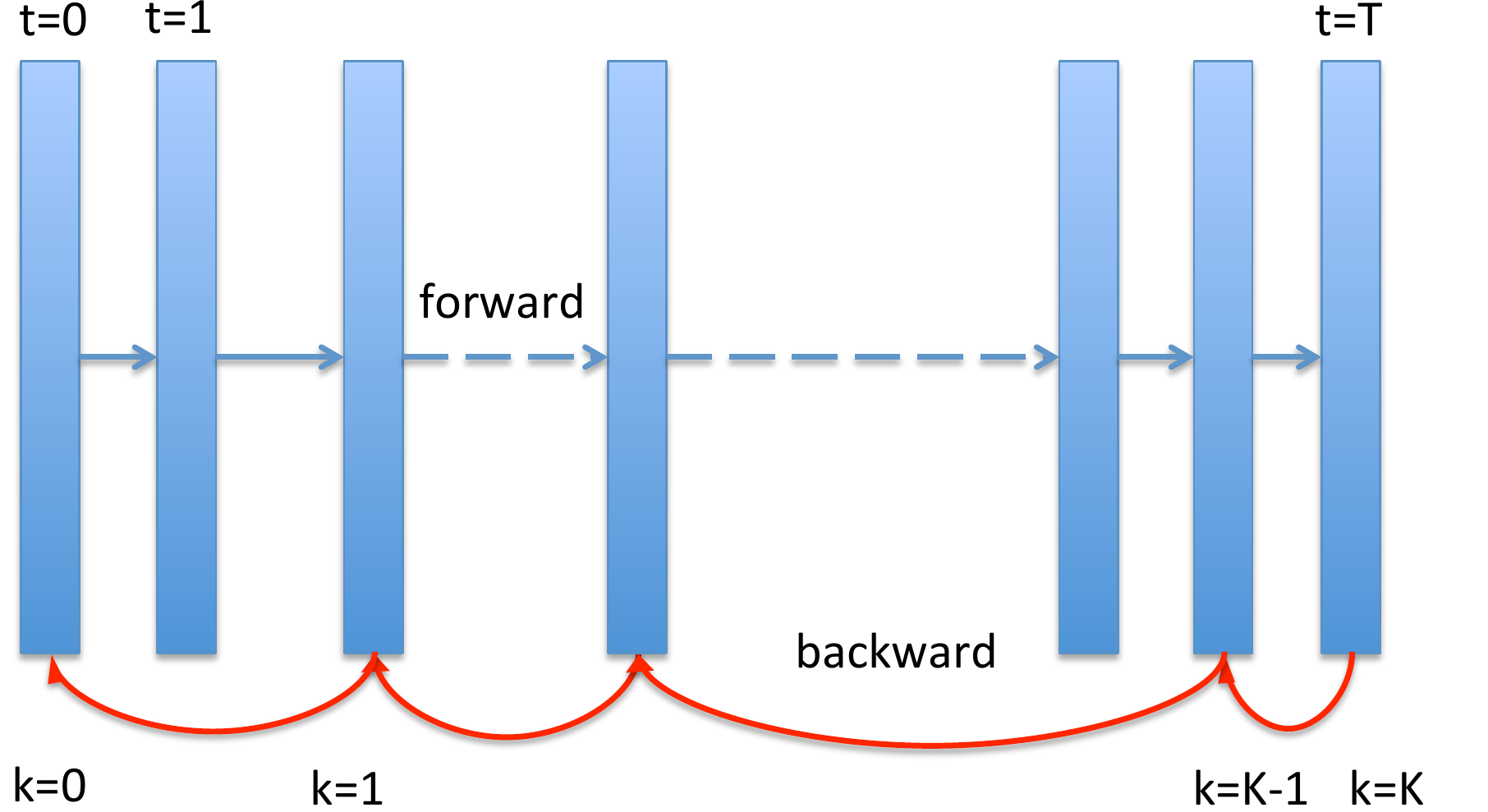}
\end{tabular}
\caption{The forward process over $T$ steps and the reverse process with shortcut MCMC sampling (red line).}
\label{Fig:sc_mcmc}
\end{figure*}
In the following part, we will focus on the fidelity term $\log p(\mathbf{x}_0 | z)$, and we want the data generated from the latent space match the original data distribution.

\subsection{Shortcut MCMC sampling}
The fidelity term  $\mathbb{E}_{q} \log p(\mathbf{x}_0 | z)$ is hard to optimize, because its complexity is determined by the depth of the generative model and its neural nets architecture. In the training stage, we always set a large $T$, such as $T=1000$. We use the forward posterior to match $p(\mathbf{x}_{t-1}|\mathbf{x}_t)$. In other words, we have $\mathcal{N} (\mathbf{x}_{t-1}; \mu_{\theta}(\mathbf{x}_t, t), \sigma_\theta(\mathbf{x}_t, t ))$ and needs to recover the data step by step. 

For any time step $s$ and $t$ $\in [1, T]$ and $s<t$, we have $q(\mathbf{x}_s | \mathbf{x}_t, \mathbf{x}_0) = \mathcal{N}(\mathbf{x}_s; \boldsymbol{{\mu}_{\theta}}(\mathbf{x}_t; s, t), \sigma^2_\theta(s, t) \bf{I} )$, with mean and variance as below
\begin{align}
\boldsymbol{{\mu}_{\theta}}( \mathbf{x}_t; s, t ) &=  \frac{ \alpha_{t|s} \sigma^2_{s} }{\sigma^2_{t}} \mathbf{x}_t  +  \frac{ \alpha_s \sigma^2_{t|s}}{\sigma_t^2}\mathbf{x}_0, \quad \sigma^{2}_\theta(s, t) = \sigma^2_{t|s}\sigma^2_{s}/\sigma^2_{t} 
\end{align}
Using KL divergence, $p(\mathbf{x}_s | \mathbf{x}_t) = q(\mathbf{x}_s | \mathbf{x}_t, \mathbf{x}_0)$, and we need to replace $\mathbf{x}_0$ with $\mathbf{\hat{x}}_0 (\mathbf{x}_t, t)$ in the inference. After do some mathematical operations in Appendix B, we have the following formula
\begin{align}\label{eq:mcmc}
p(\mathbf{x}_s ) = \alpha_s \mathbf{x}_0  + \sigma_s \epsilon
\end{align}
Thus, we can sample $\mathbf{x}_s$ at any time step $s$. In the best scenario, the marginal distribution $p(\mathbf{x}_t)$ from the reverse process matches the forward one $q(\mathbf{x}_t)$. 
Since we have $p(\mathbf{x}_t) \sim q(\mathbf{x}_t)$, we approximate $p(\mathbf{x}_t)$ with the same formula in Eq. \ref{eq:fw} and we can sample $\mathbf{x}_s$ from the constructed $\hat{\mathbf{x}}_0$. Since the latent variable $z=(\mathbf{x}_1,..., \mathbf{x}_T)$, it will be time-consuming. To speed up the inference, we can skip steps to produce data while using MCMC sampling. Specifically, we random sample $K$ time steps $\{t_1,..,t_K\}$ from $[1, T]$. Then we use the prediction $\hat{\mathbf{x}}_{t_{k}}$ to get the next sample $\hat{\mathbf{x}}_{t_{k-1}}$ according to the equation above. Thus we have the fidelity loss
\begin{align}\label{eq:floss}
\mathbb{E}_{q} \log p(\mathbf{x}_0 | z) = || \mathbf{x}_0 - \hat{ \mathbf{x}_0} ||^2
\end{align}
where $\hat{x_0}$ is predicted from the shortcut MCMC sampling. By minimizing this loss, we add the global constraint to the deep denoise models, and further improve the data approximation quality. 

\subsection{Algorithm}
We summarize our approach in Algorithm. \ref{alg:algd2q}. Compared to DDPMs and VDMs, we add the fidelity term which imposes a global constraint to our generated samples and use shortcut MCMC sampling to speed up the inference. 

In the inference stage, we just sample $\epsilon \sim \mathcal{N}(0, \mathbf{I} )$, then we sample K time steps from $[1, T]$ and sample $\mathbf{x}_{t_k} \sim \alpha_{t_k} \hat{\mathbf{x}_0} + \sigma_{t_k} \epsilon$, where $\mathbf{\hat{x}}_0$ is predicted from the denoise neural network in the previous $t_{k-1}$. Thus, our method has the potential to speed up inference at least an order of magnitude fast.
\begin{algorithm}[tb]
\caption{Training}
\label{alg:algd2q}
\begin{algorithmic}
\STATE Initialize denoise neural networks and its parameters
\FOR{epoch = 1 {\bfseries to} $N$}
\STATE $x_0 \sim q(x_0)$
\STATE sample $t \sim$ Uniform(1,..., T) 
\STATE take step to minimize $|| \epsilon - \hat{\epsilon}_\theta (x_t, t) ||^2$, where $x_t = \alpha_t x_0 + \sigma_t \epsilon$  
\STATE random sample K steps (not need to be equal distance), $t_0, t_1,..., t_K \sim$  Uniform(1,..., T)
\FOR{k = K {\bfseries to} $1$}
\STATE predict $\hat{x_0} = (x_{t_k} - \sigma_{t_k} \hat{\epsilon}_\theta (x_{t_k}, t) )/\alpha_{t_k}$
\STATE update $x_{t_{k-1}} \sim \alpha_{t_{k-1}} \hat{x_0} + \sigma_{t_{k-1}} \epsilon $
\ENDFOR
\STATE take gradient step to minimize $|| x_0 - \hat{x}_0||^2$
\ENDFOR
\STATE Return model and parameters.
\end{algorithmic}
\end{algorithm}

\section{Experimental results}
We did initial experiments on synthetic dataset. In this experiment, we create the swirl dataset with 1024 points, shown in Fig \ref{Fig:noise}. As for the model architecture, we use 3 layer MLP, with Fourier feature expansion as the inputs. We set $K=10$ for all the training in all the experiments below. 

\begin{figure*}[h!]\center
\begin{tabular}{ccc}
\includegraphics[trim=13mm 8.0mm 20mm 13.5mm, clip, width=5.cm]{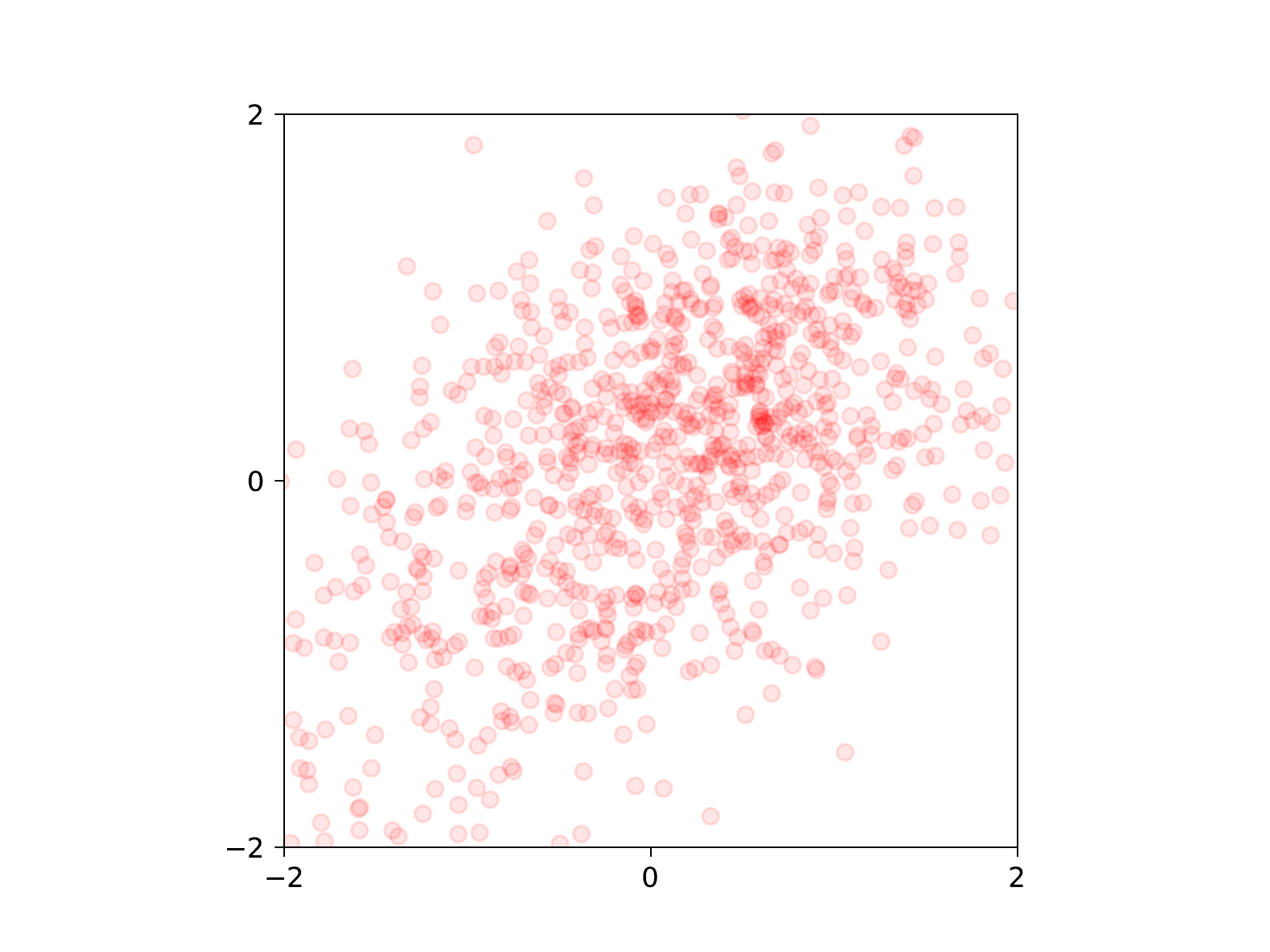} &
\includegraphics[trim=13mm 8.0mm 20mm 13.5mm, clip, width=5.cm]{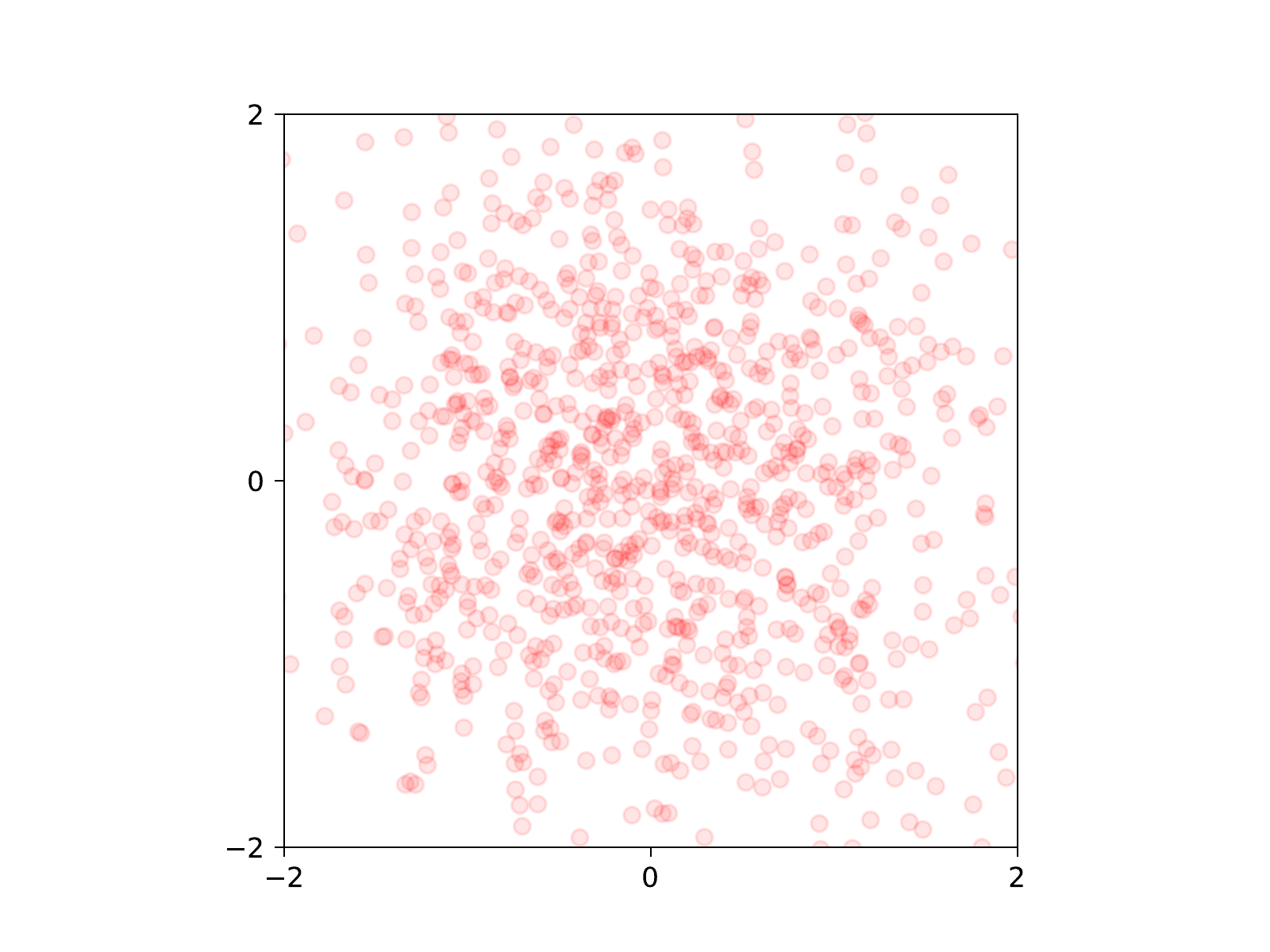} & k=0\\
\includegraphics[trim=13mm 8.0mm 20mm 13.5mm, clip, width=5.cm]{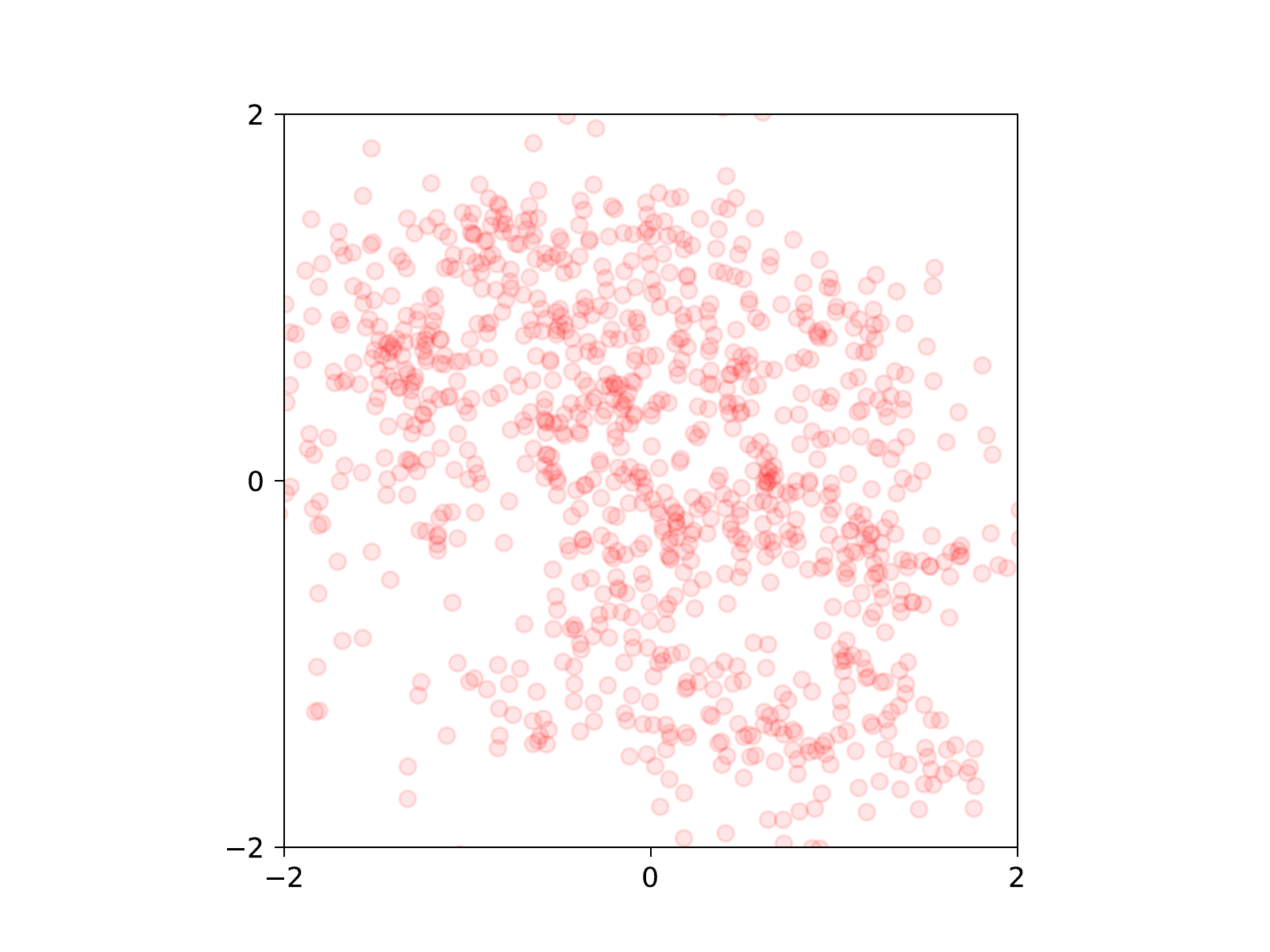} &
\includegraphics[trim=13mm 8.0mm 20mm 13.5mm, clip, width=5.cm]{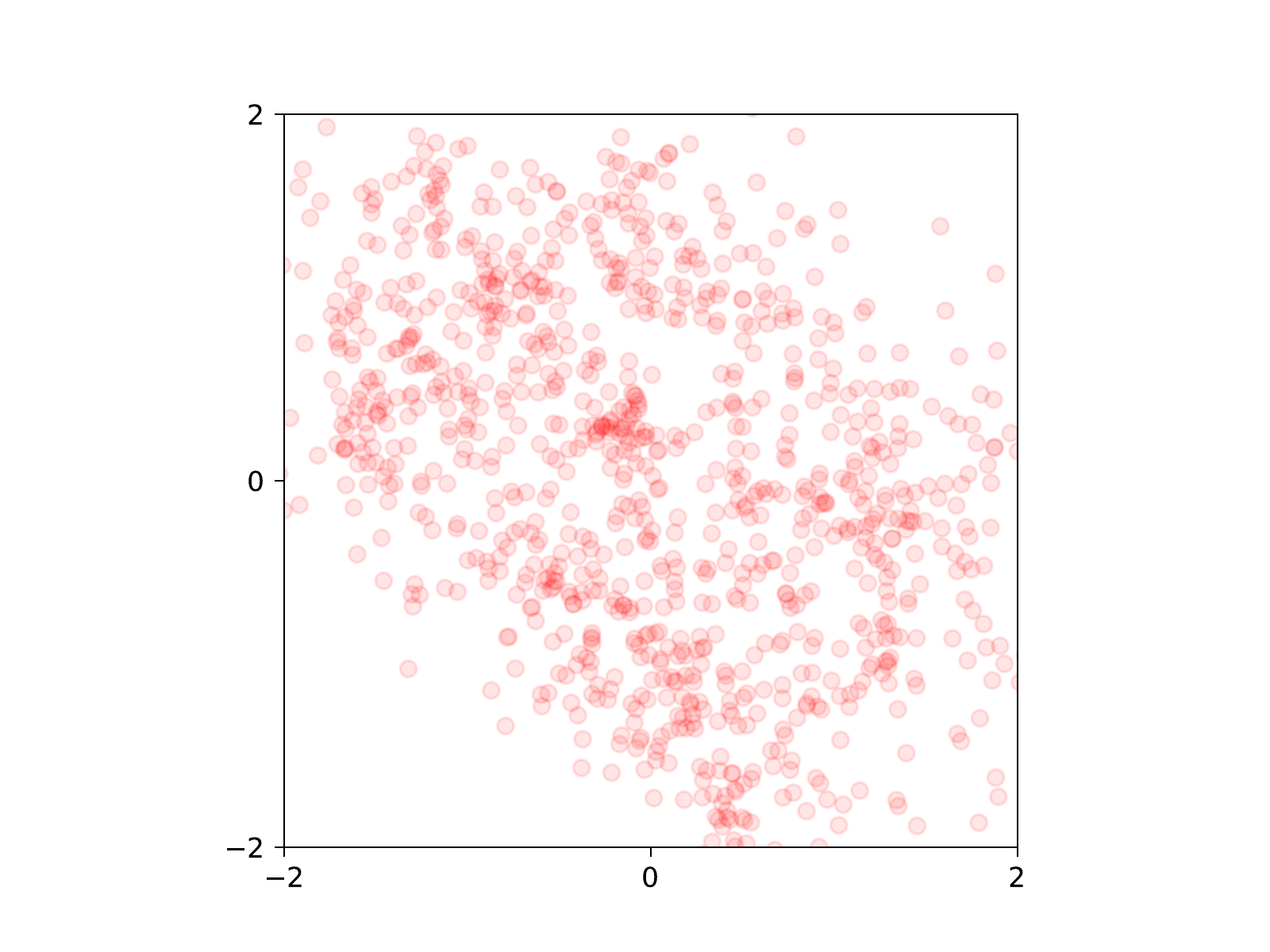} & k=2 \\ 
\includegraphics[trim=13mm 8.0mm 20mm 13.5mm, clip, width=5.cm]{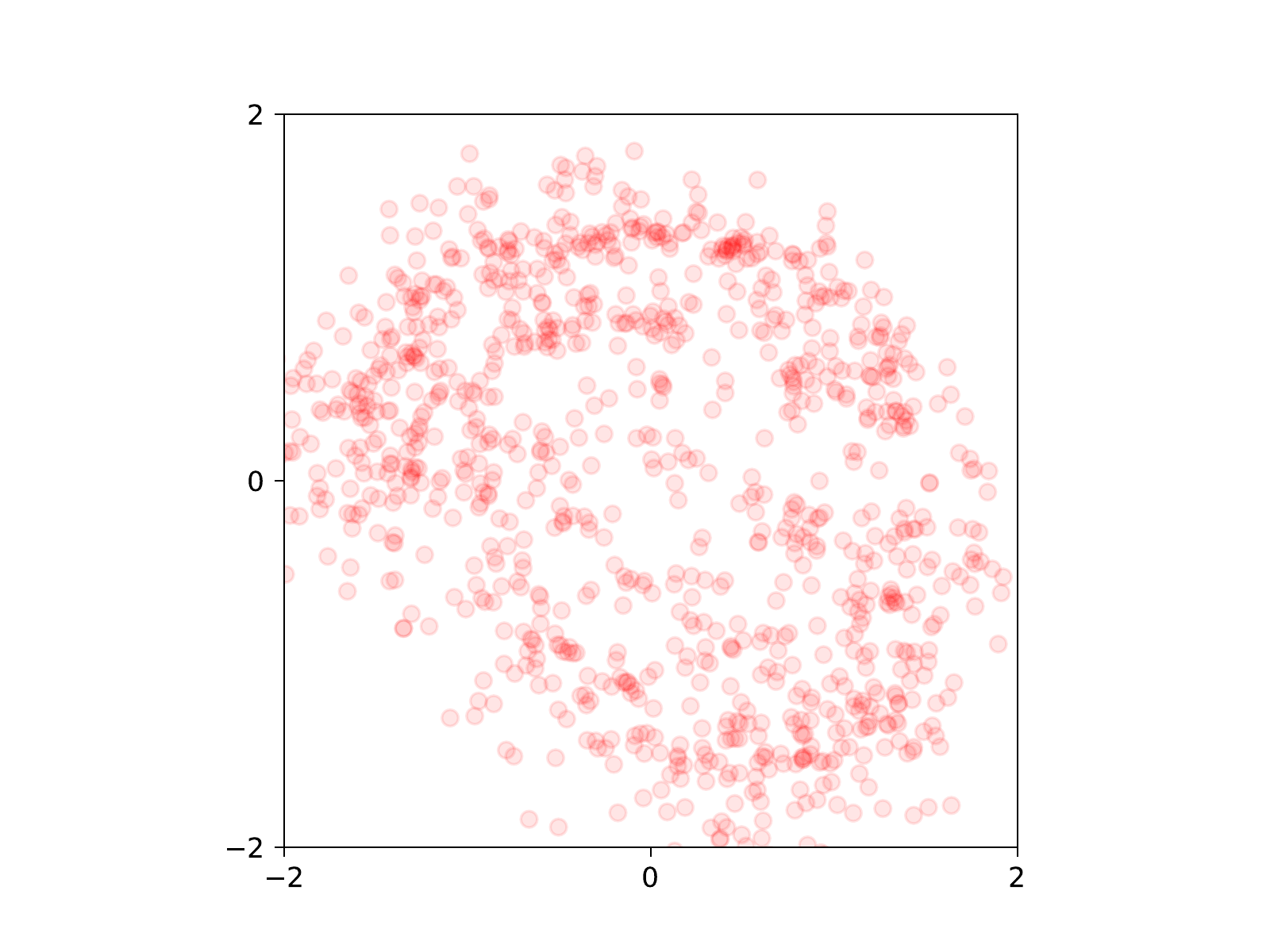} &
\includegraphics[trim=13mm 8.0mm 20mm 13.5mm, clip, width=5.cm]{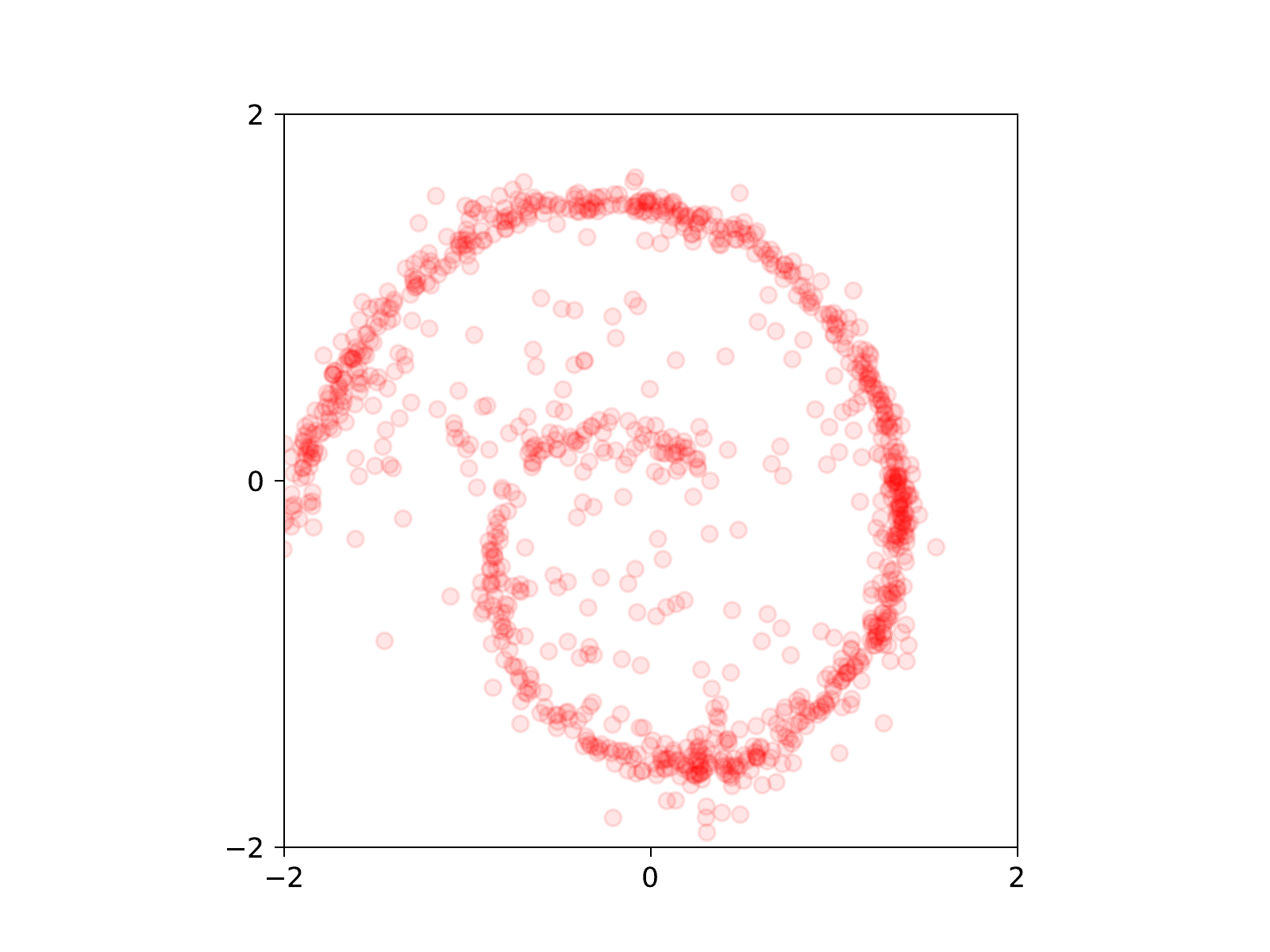} & k=4\\
\includegraphics[trim=13mm 8.0mm 20mm 13.5mm, clip, width=5.cm]{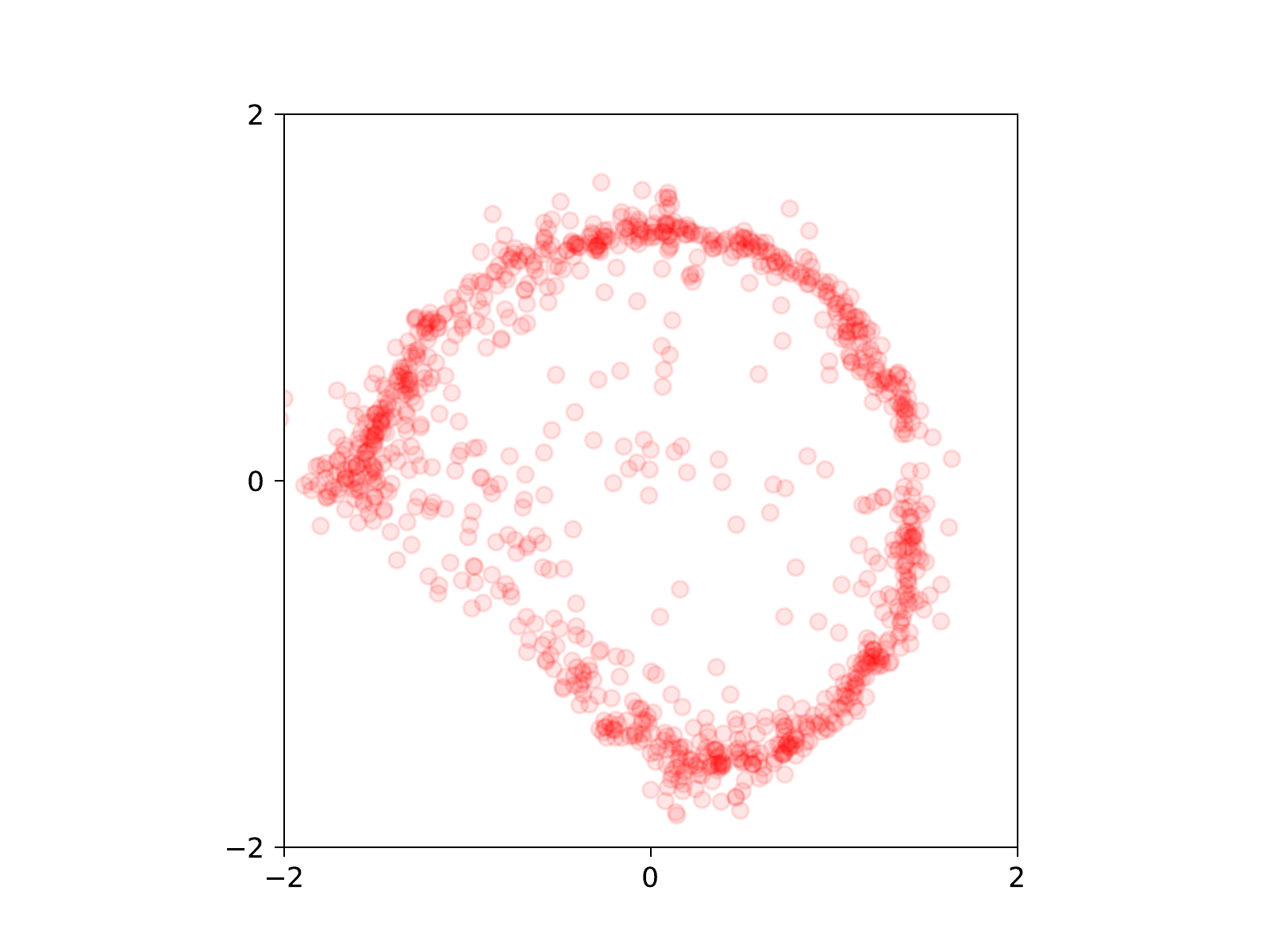} &
\includegraphics[trim=13mm 8.0mm 20mm 13.5mm, clip, width=5.cm]{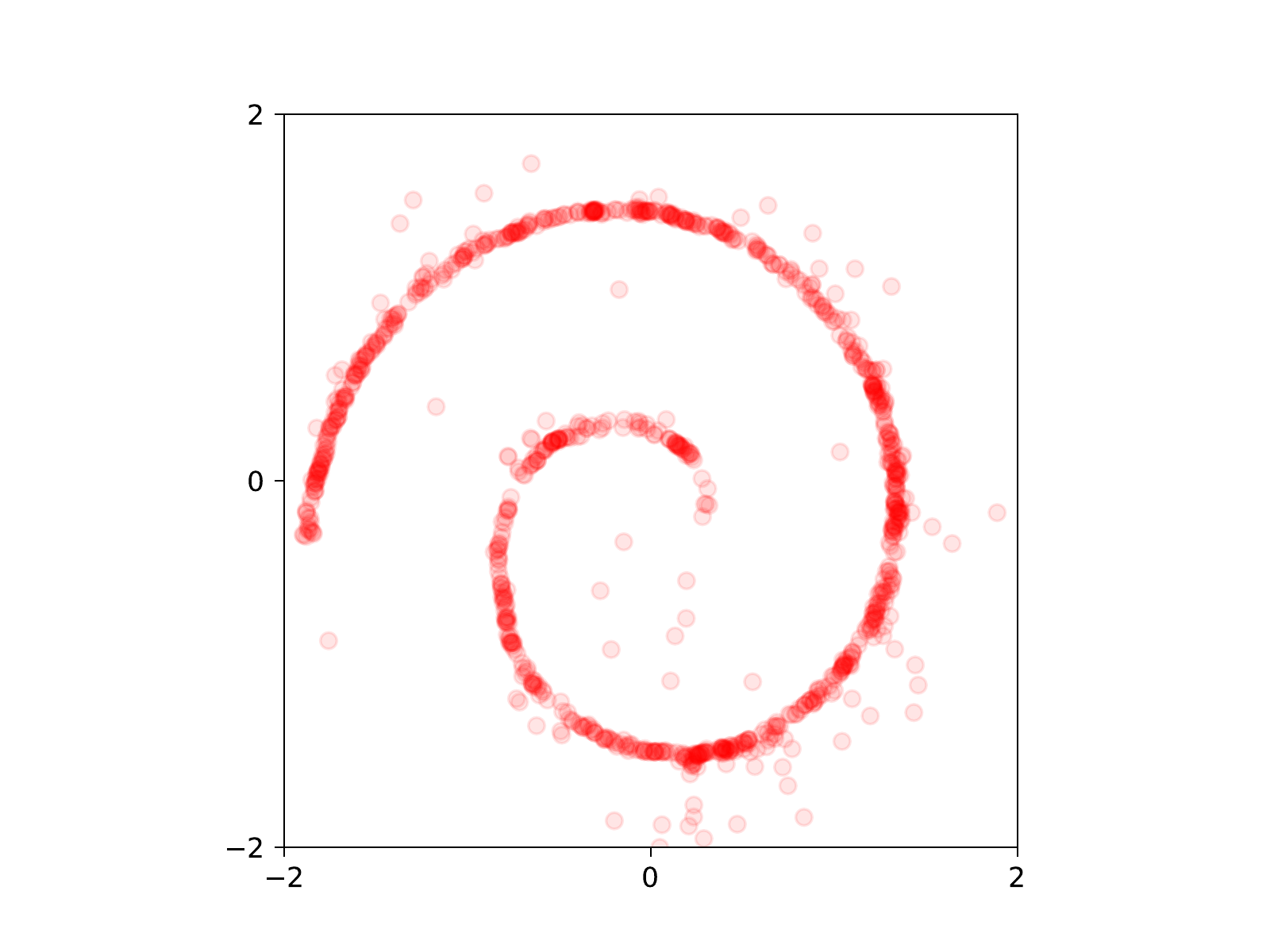} & k=7\\ 
\includegraphics[trim=13mm 8.0mm 20mm 13.5mm, clip, width=5.cm]{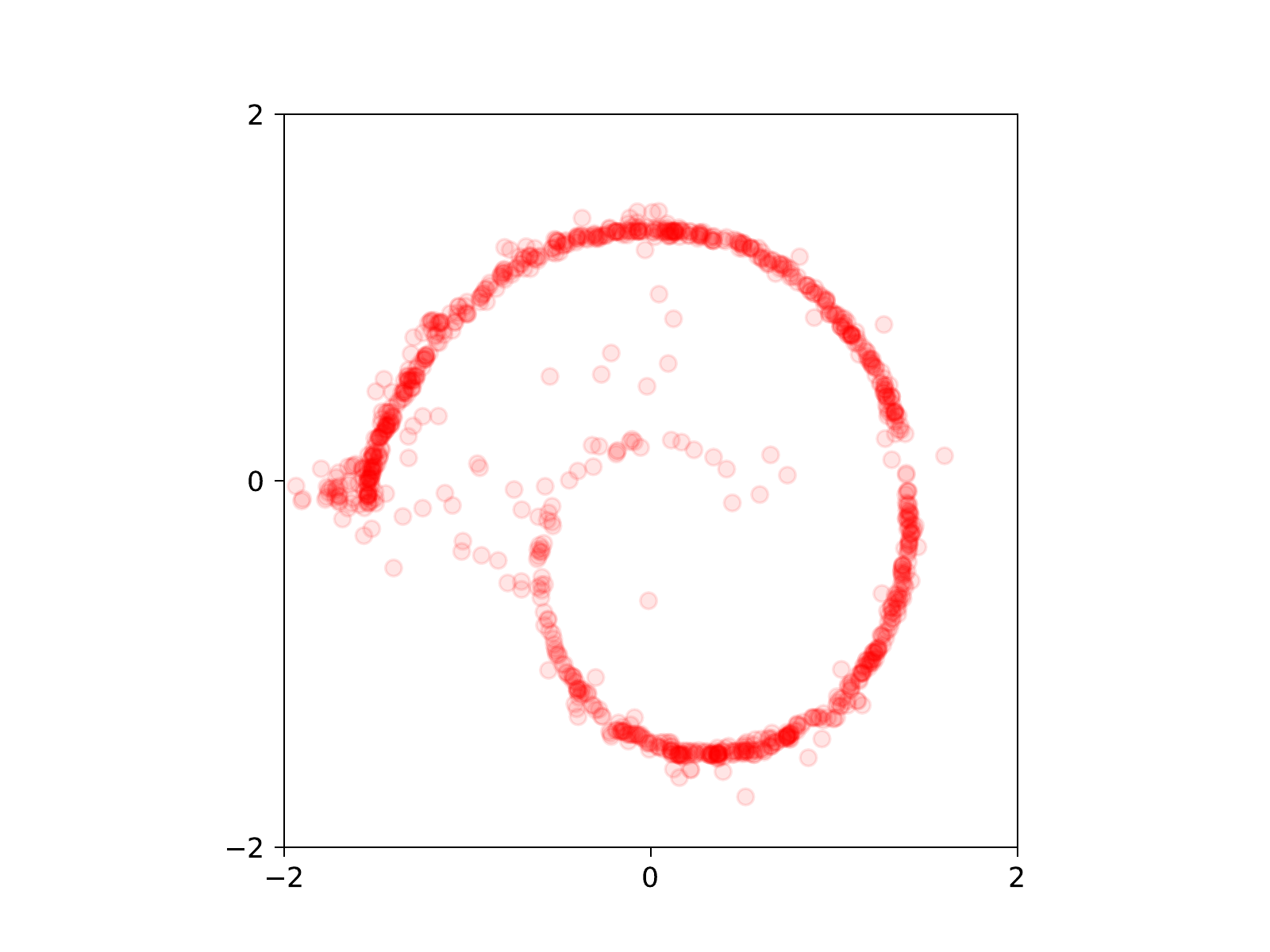} &
\includegraphics[trim=13mm 8.0mm 20mm 13.5mm, clip, width=5.cm]{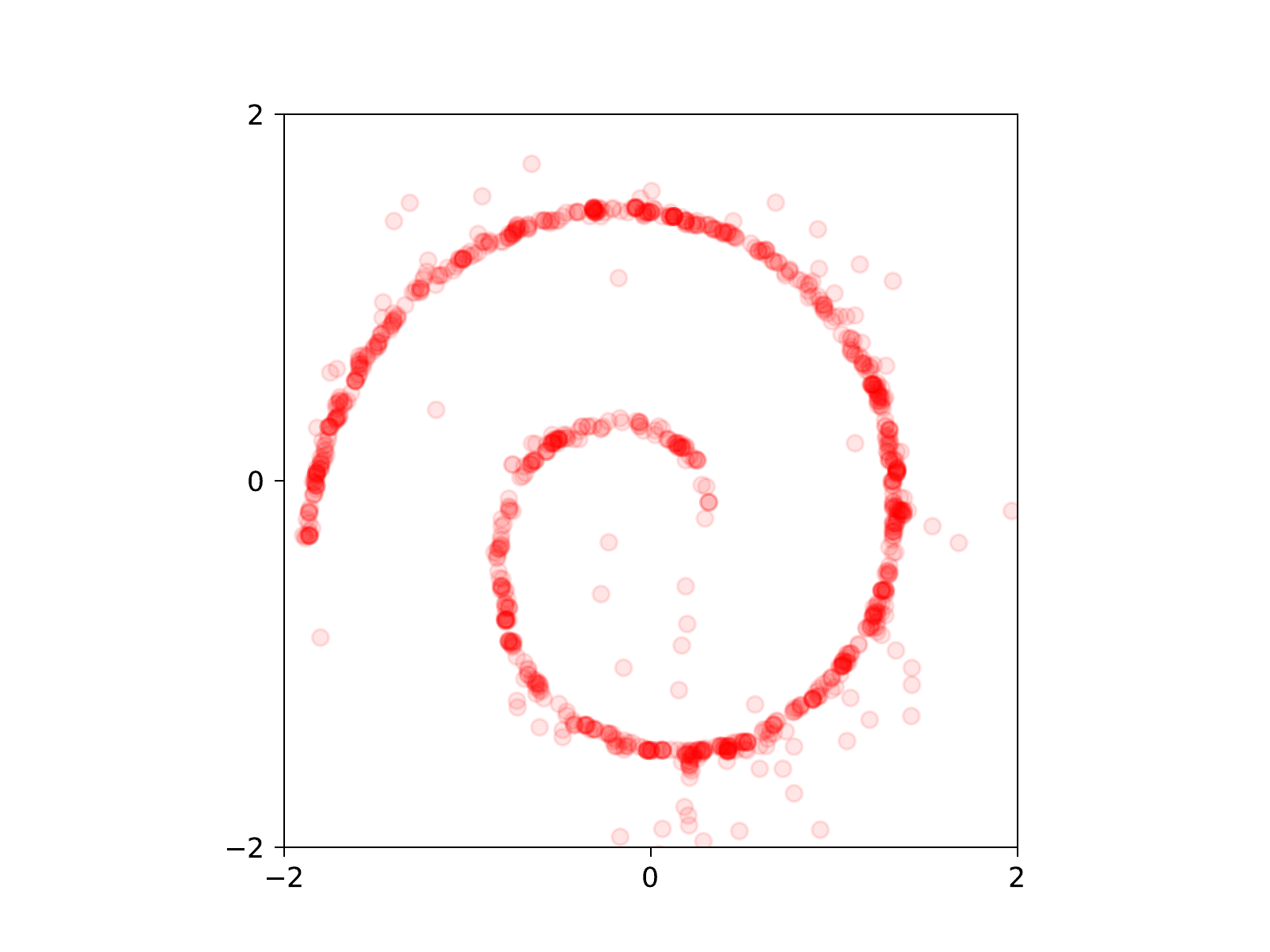} & k = 10\\
(a) VDMs & (b) ours with MCMC sampling & the time step
\end{tabular}
\caption{The left column is from VDMs\cite{Kingma21}, the right column is from our approach. We use $T=200$ in the inference stage, and K=10 to sample 10 time steps. Then we compare the corresponding 5 generated images between VDMs and our method. }
\label{Fig:k10}
\end{figure*}

In the first experiment in Fig \ref{Fig:k10}, we train the model with the shortcut MCMC sample. In the inference stage, we set $T=200$ and sample $K=10$ time steps, then we generate our results with only 10 steps inference. The result in Fig \ref{Fig:k10} shows that our approach not only converge fast, but also reconstruct better results. 

\begin{figure*}[h!]\center
\begin{tabular}{ccc}
\includegraphics[trim=13mm 8.0mm 20mm 13.5mm, clip, width=5.2cm]{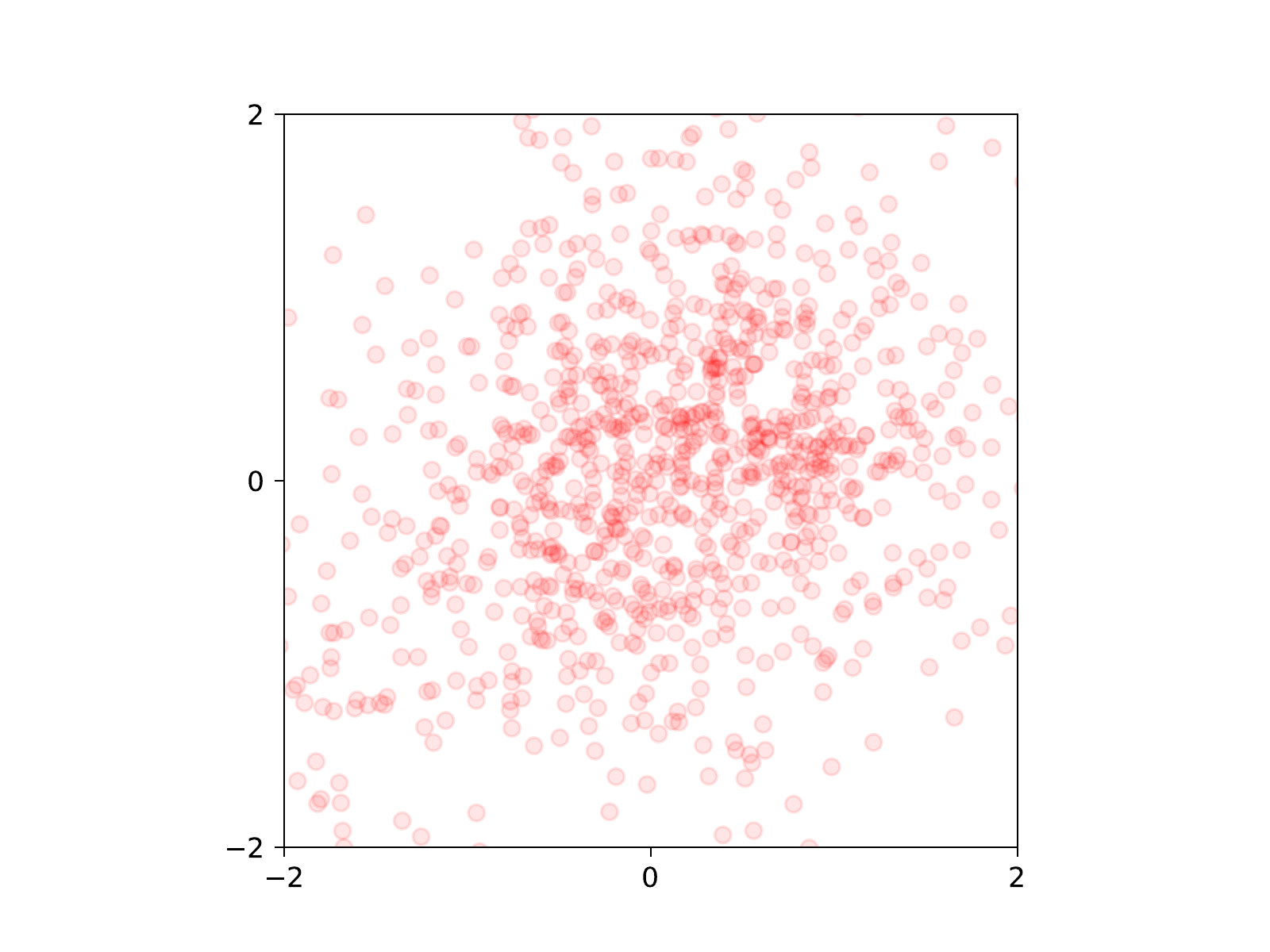} &
\includegraphics[trim=13mm 8.0mm 20mm 13.5mm, clip, width=5.2cm]{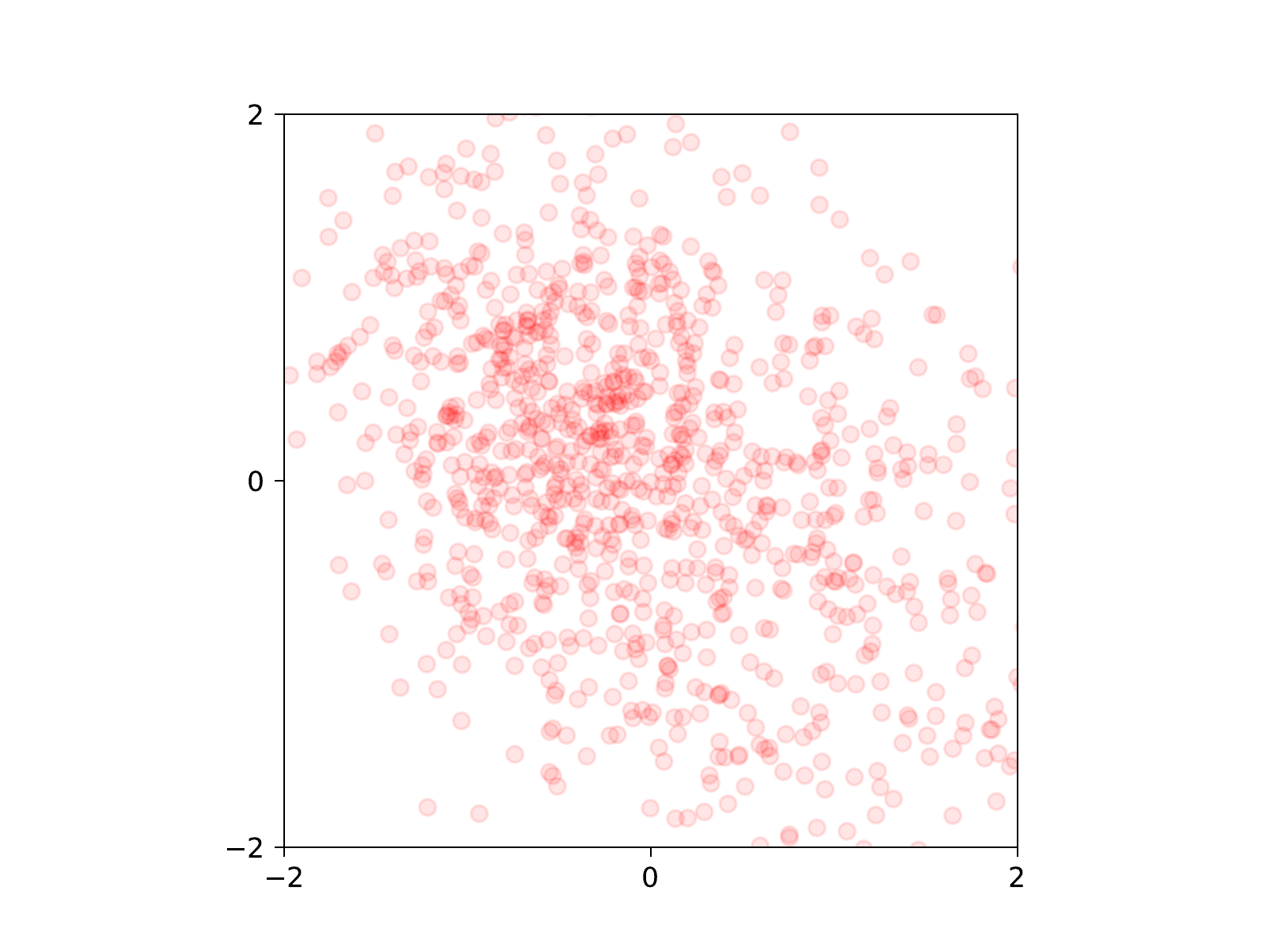} & k =20\\ 
\includegraphics[trim=13mm 8.0mm 20mm 13.5mm, clip, width=5.2cm]{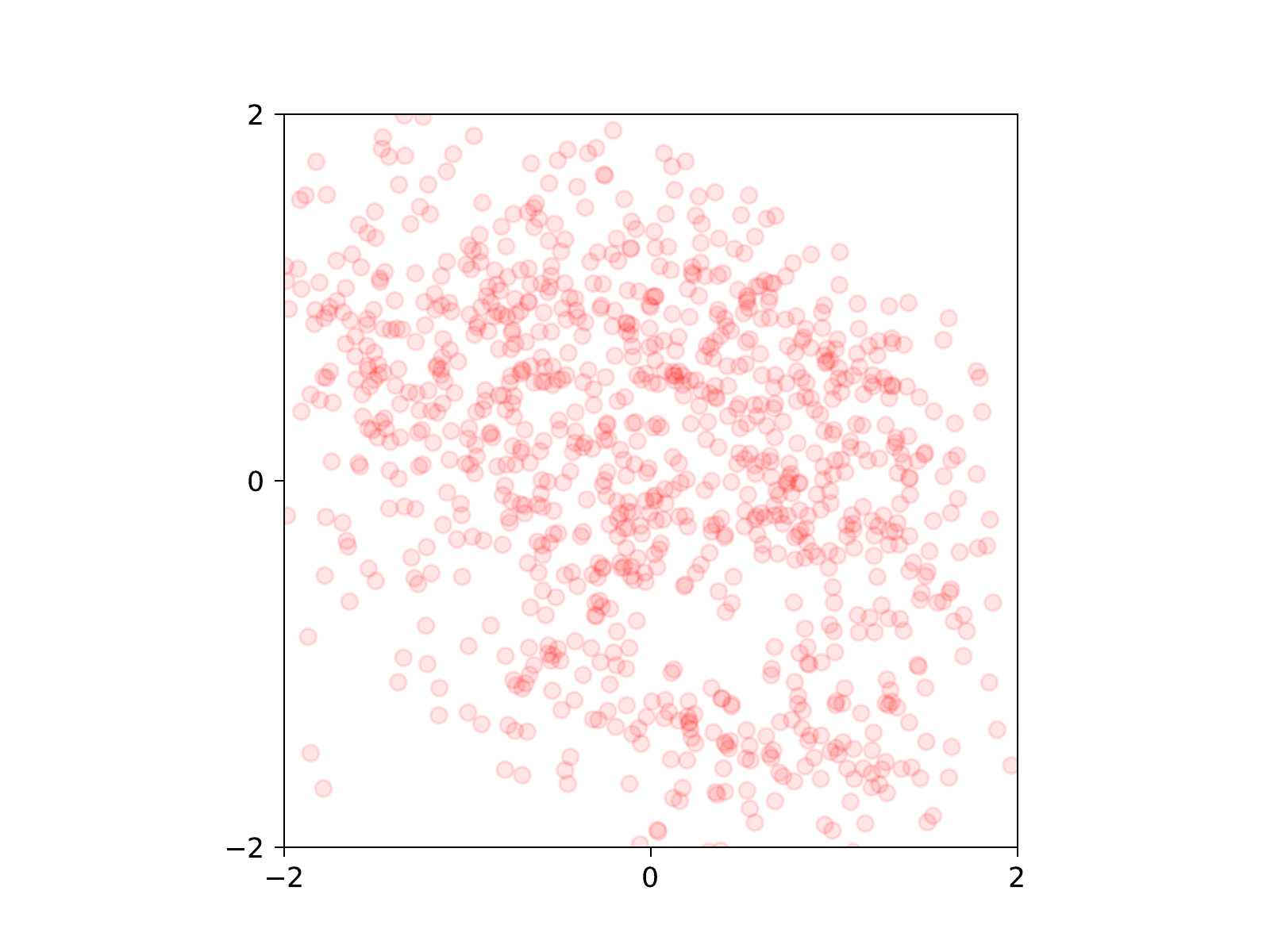} &
\includegraphics[trim=13mm 8.0mm 20mm 13.5mm, clip, width=5.2cm]{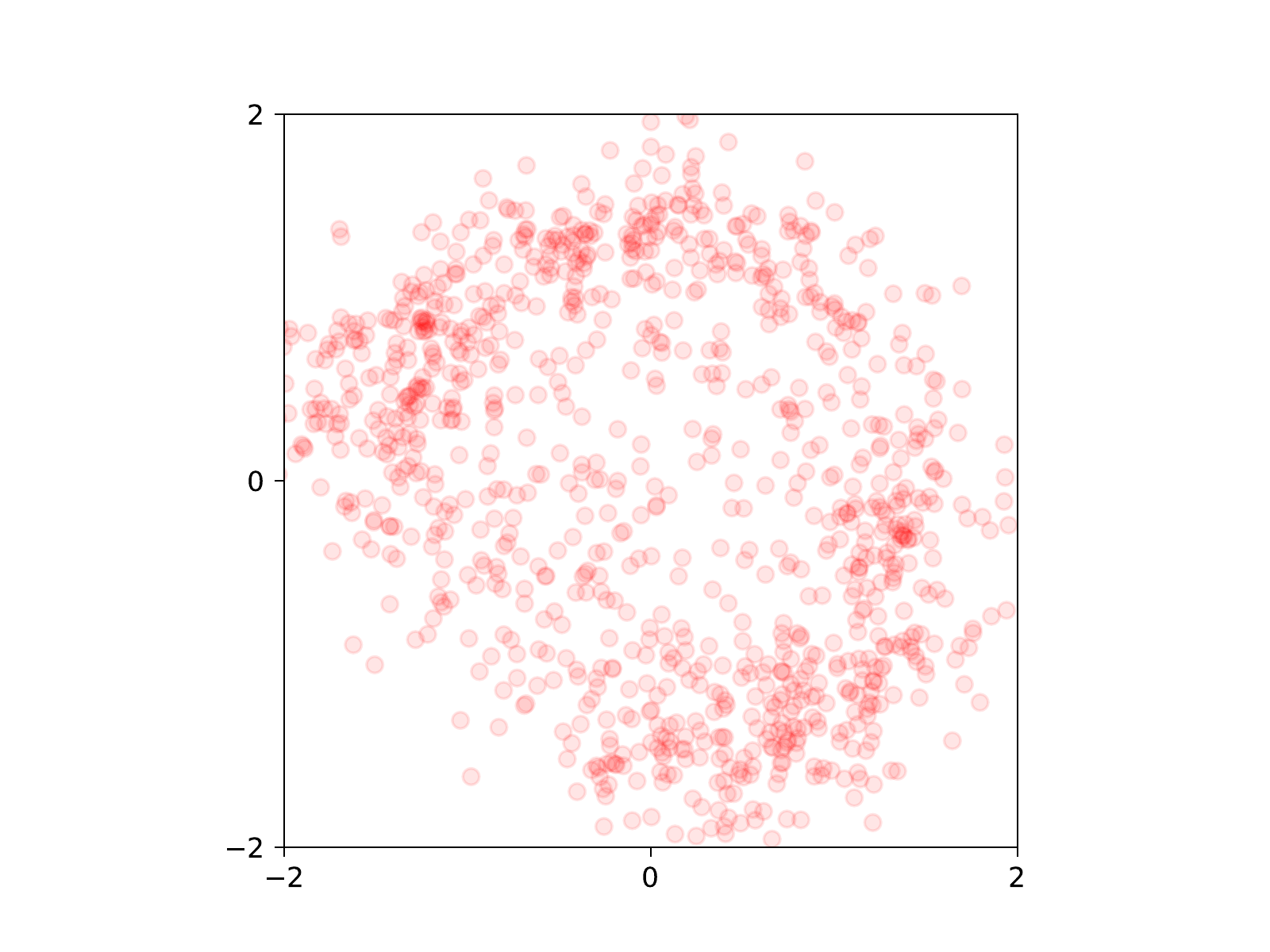} & k=60\\ 
\includegraphics[trim=13mm 8.0mm 20mm 13.5mm, clip, width=5.2cm]{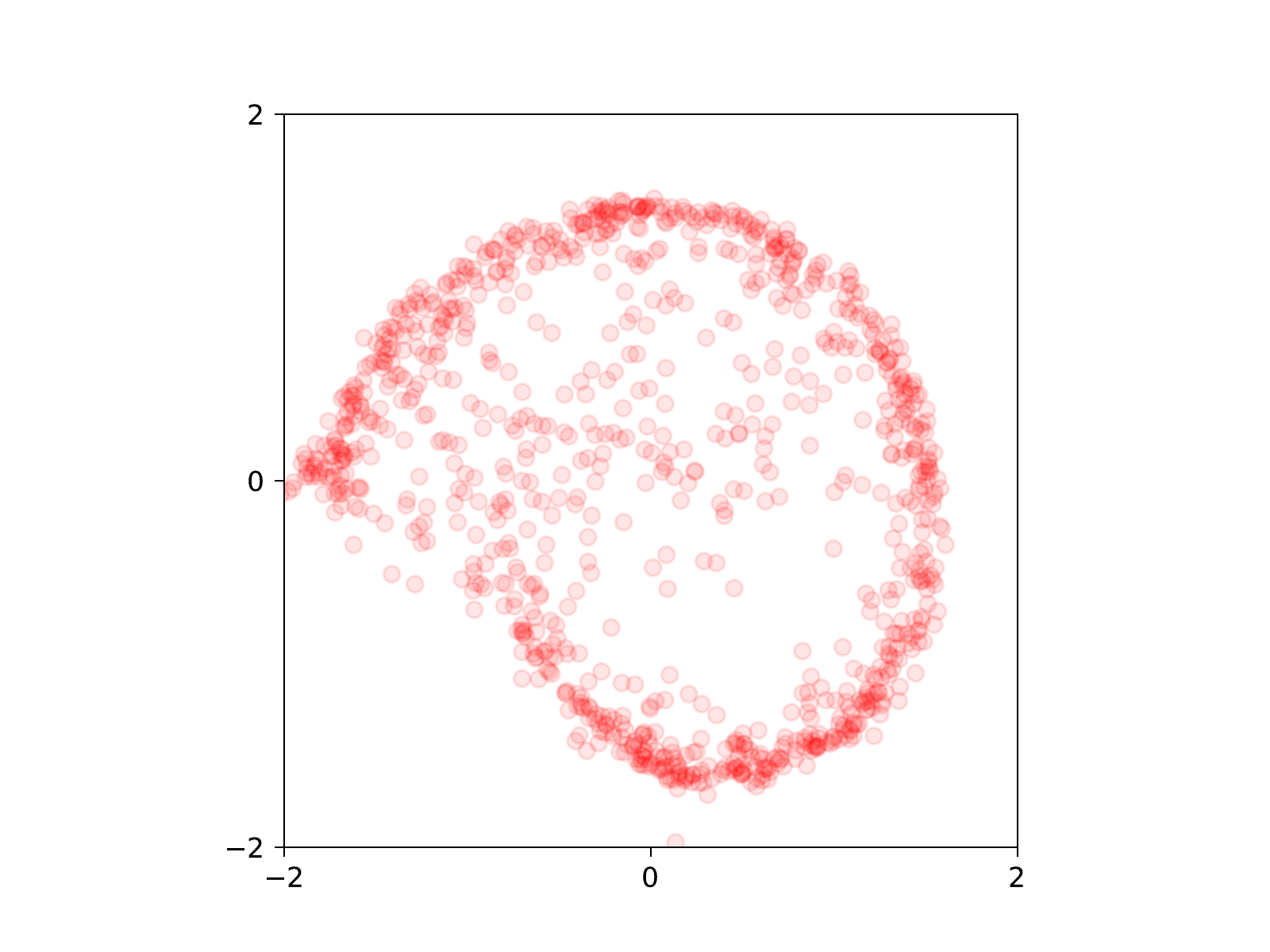} &
\includegraphics[trim=13mm 8.0mm 20mm 13.5mm, clip, width=5.2cm]{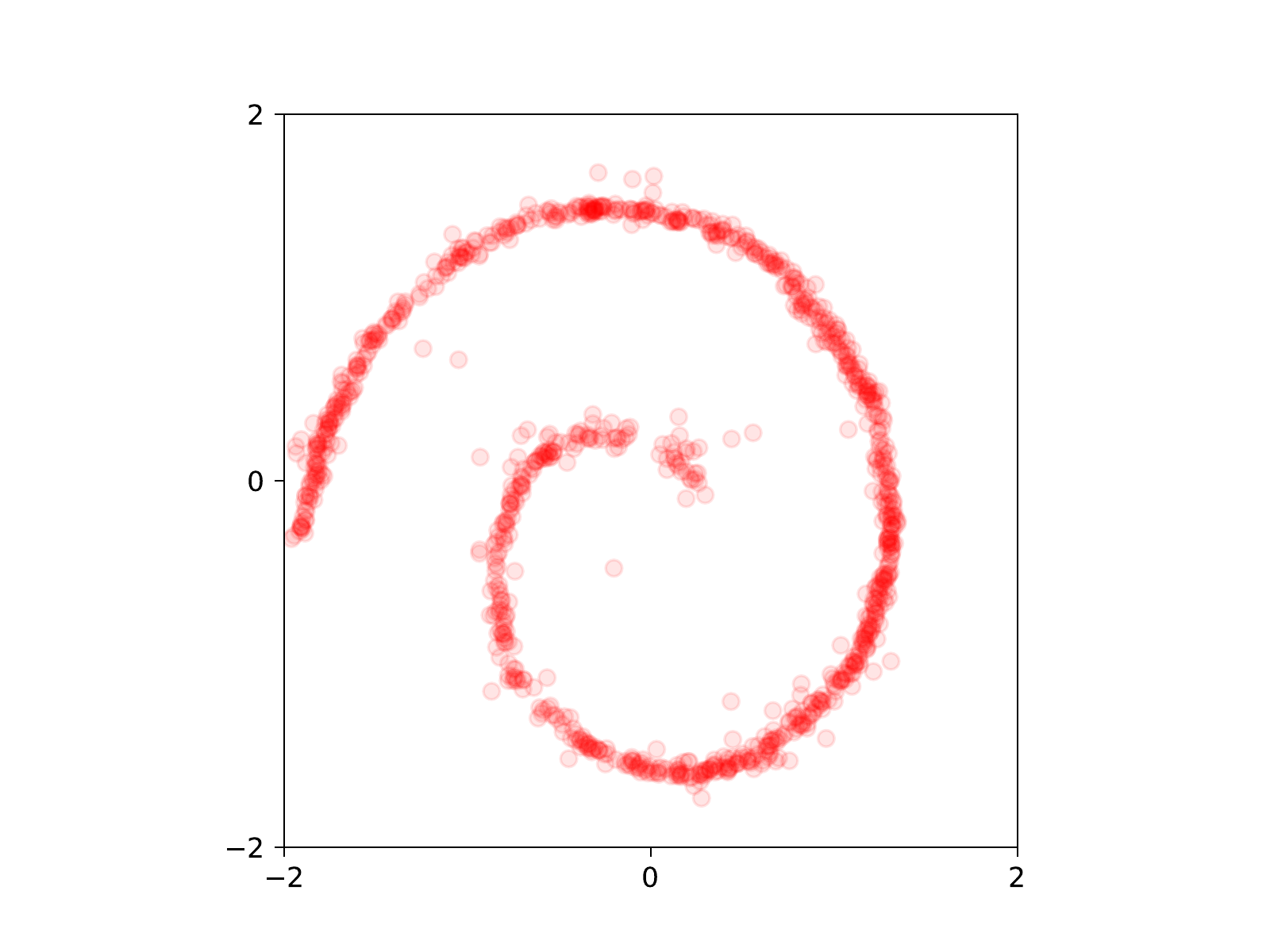} & k = 100\\ 
\includegraphics[trim=13mm 8.0mm 20mm 13.5mm, clip, width=5.2cm]{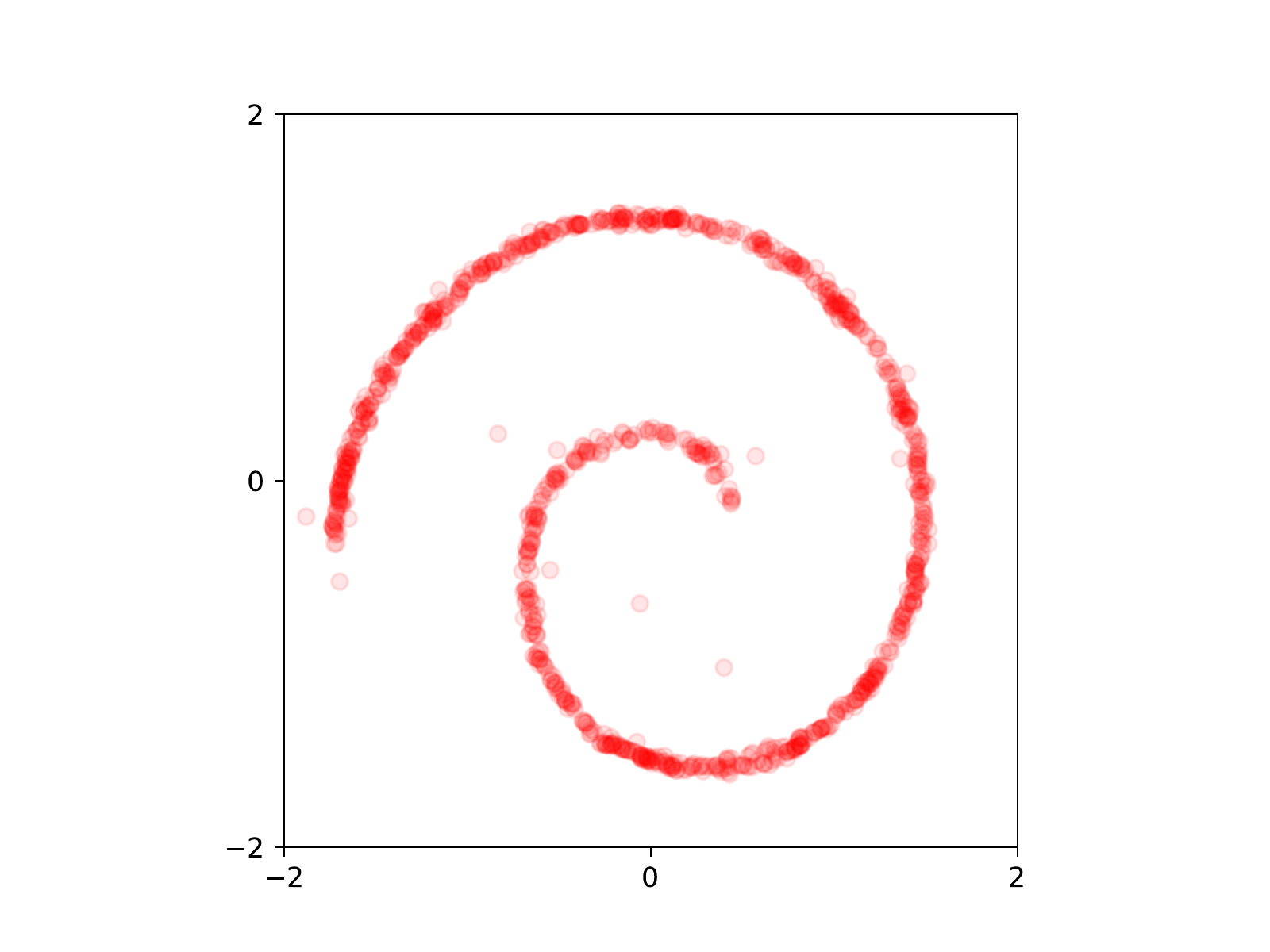} &
\includegraphics[trim=13mm 8.0mm 20mm 13.5mm, clip, width=5.2cm]{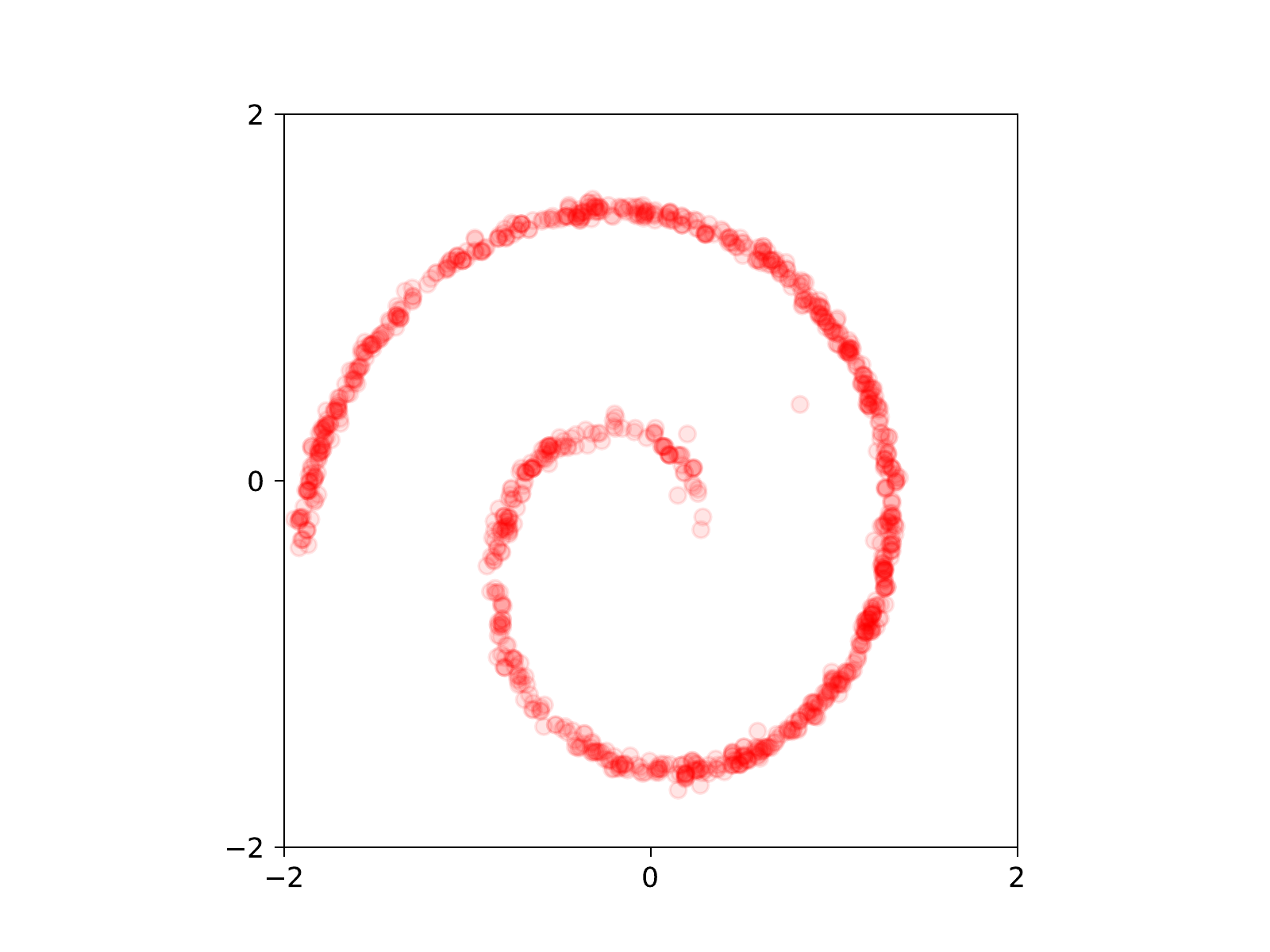} & k =160\\ 
\includegraphics[trim=13mm 8.0mm 20mm 13.5mm, clip, width=5.2cm]{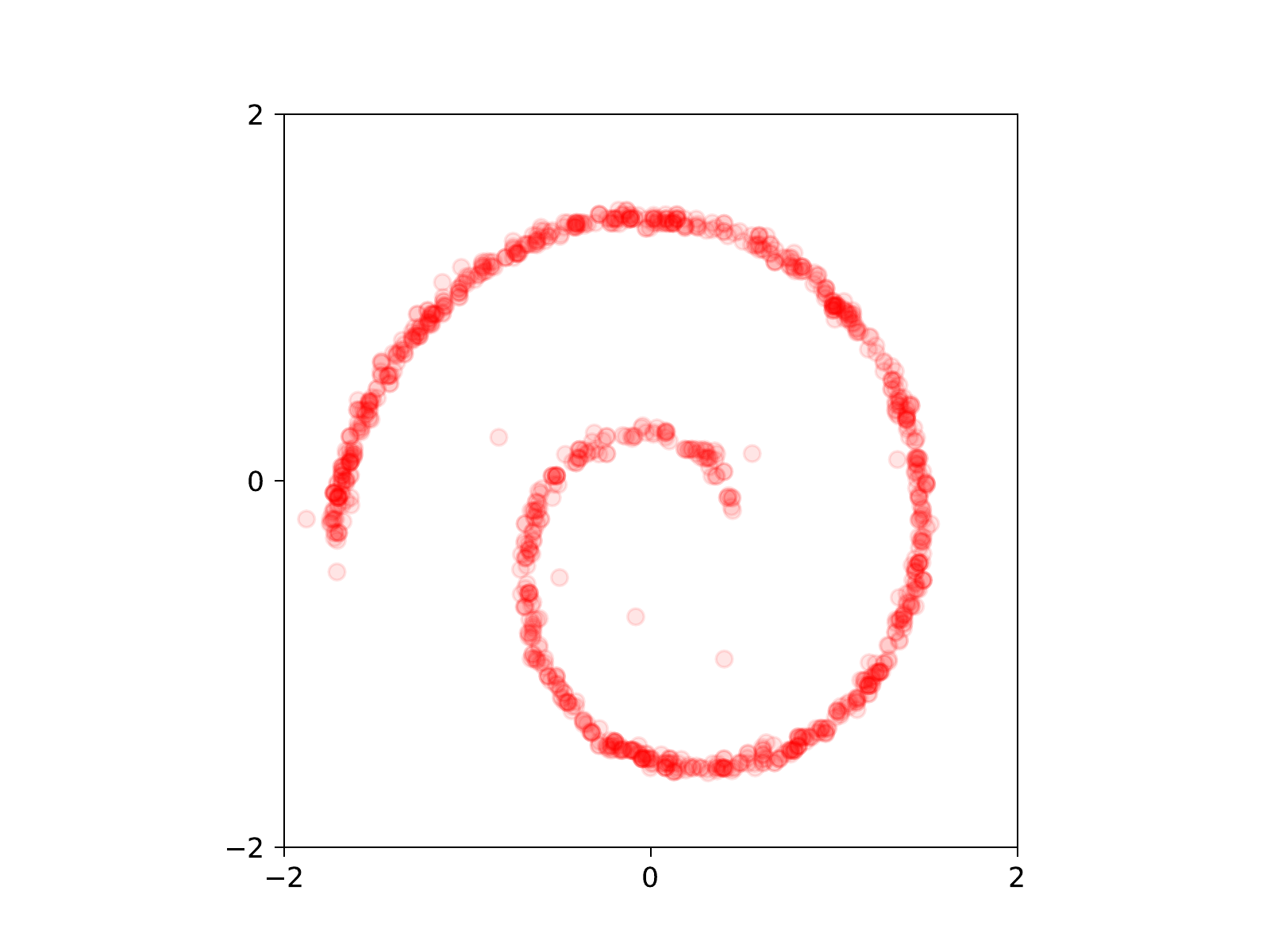} &
\includegraphics[trim=13mm 8.0mm 20mm 13.5mm, clip, width=5.2cm]{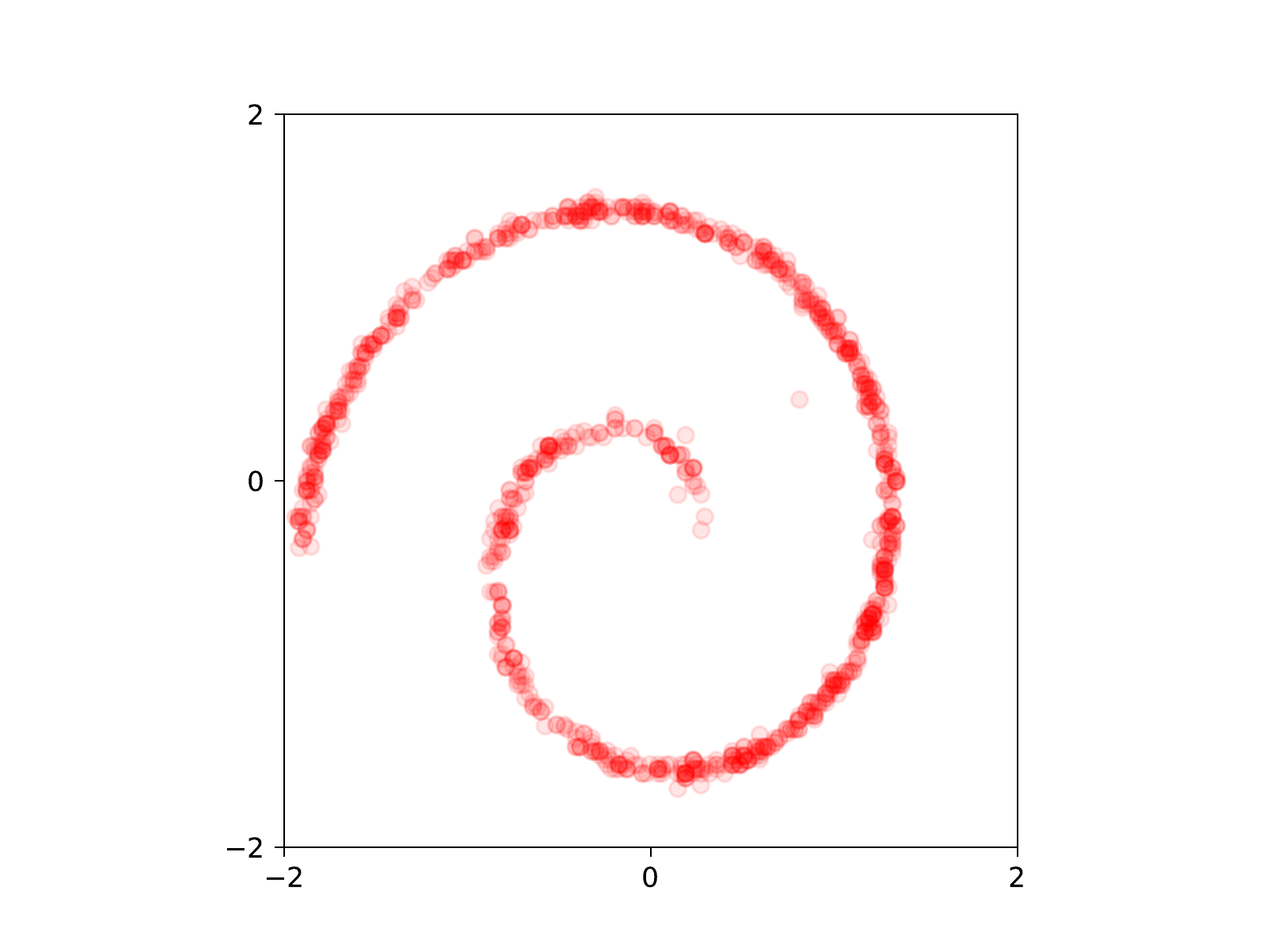} & k=200\\
(a) VDMs & (b) ours with MCMC sampling & the time step
\end{tabular}
\caption{We use $T=200$ in the inference stage, and K=200 for the full time steps comparison. We can see our method can generate very good samples and converge fast then VDMs.}
\label{Fig:t200}
\end{figure*}

In the second experiments, we train with $K=10$, and in the inference we set $K$ the same value as $T$, $K=T=200$ for step by step comparison. It indicates that with the same time steps, our approach converge fast and yield better results in Fig \ref{Fig:t200}. For example, our approach recover the data well at $K=100$.

\section{Conclusion}
In this paper, we propose a fast approach for diffusion models in the inference stage. To this end, we add a fidelity term as the global constraint over the diffusion models, and present a shortcut MCMC sampling method to speed up the inference. The experiments show promising results on both data quality and fast inference time. 
%
%
\section{Appendix}
\appendix 
\section{}
The maximum likelihood $x_0$ is
\begin{align}\label{eq:ml}
\log p(x_0) &= \log \int_z p(x_0, z) = \log \int_z p(x_0, z) \frac{q(z|x)}{q(z|x)} = \log \int_z q(z|x)  \frac{p(x_0, z)}{q(z|x)}  \nonumber \\
&\ge \int q(z|x) \log \frac{p(x, z)}{q(z|x)} = E_{q(z|x)} [ \log \frac{p(x,z)}{q(z|x)}  ] = E_{q(z|x)} [ \log \frac{p(x|z)p(z)}{q(z|x)}  ]  \nonumber \\
&= E_{q(z|x)} [ \log p(x|z)] - E_{q(z|x)} [\log \frac{p(z)}{q(z|x)} ]  
\end{align}
where we assume the latent $z = (x_1, x_2, ..., x_T)$. Overall, we want to maximize the variational lower bound. The first term is reconstruction loss, which is our fidelity term in the paper. The second term is the KL divergence between $p(z)$ and $q(z|x)$, which we want to minimize.

As for the second term we can do some decomposition to get KL divergence between $p(x_s| x_t)$ and $q(x_s|x_t, x_0)$ in the following analysis: 
\begin{align}\label{eq:kl_loss}
&\mathbb{E}_{x_{0:T} \sim q(x_{0:T})} [  \log \frac{ p(x_{1:T})} {q(x_{1:T} | x_0)} ]   \nonumber \\   
= &\mathbb{E}_{x_{0:T} \sim q(x_{0:T})} [- \log q(x_{1:T} | x_0) + \log p(x_{1:T}) ] \nonumber \\
= &\mathbb{E}_{x_{0:T} \sim q(x_{0:T})} \bigg[  -\log [ q(x_T | x_0) \prod_{t=2}^T q(x_{t-1} | x_{t}, x_0) ] + \log [ p(x_T)\prod_{t=2}^T p(x_{t-1} | x_t) ]  \bigg]\nonumber \\
=& - D_{KL} ( q(x_T |  x_0) ||p(x_T) ) - \sum_{t=2}^T D_{KL} ( q(x_{t-1} | x_{t}, x_0) || \log p(x_{t-1} | x_t) )
\end{align}

\section{}

\begin{align}\label{eq:qdloss}
p(x_s | x_t) = q(x_s|x_t, x= \hat{x}_\theta (z_t; t) )
\end{align}
Since the reverse process is also Gaussian, we then have
\begin{align}\label{eq:qdloss}
p(x_s | x_t) = \mathcal{N}(x_s; \boldsymbol{{\mu}_{\theta}}(x_t; s, t), \sigma^2_Q(s, t) \bf{I} )
\end{align}

\begin{align}\label{eq:qdloss}
\boldsymbol{{\mu}_{\theta}}(x_t; s, t) &=  \frac{ \alpha_{t|s} \sigma^2_{s} }{\sigma^2_{t}} x_t + \frac{ \alpha_s \sigma^2_{t|s}}{\sigma_t^2}\mathbf {\hat{x}_{\theta}}(x_t; t) \nonumber \\
& = \frac{1}{\alpha_{t|s}} x_t   -  \frac{ \sigma^2_{t|s} }{ \alpha_{t|s} \sigma_t } \mathbf{\hat{\epsilon}_{\theta}}(x_t; t) \nonumber \\
& = \frac{1}{\alpha_{t|s}}(\alpha_t \mathbf{x} + \sigma_t \mathbf{\epsilon} ) -  \frac{ \sigma^2_{t|s} }{ \alpha_{t|s} \sigma_t } \mathbf{\hat{\epsilon}_{\theta}}(x_t; t) \nonumber \\
& = \alpha_s \mathbf{x} + \frac{1}{\alpha_{t|s}} (    \sigma_t \mathbf{\epsilon} -    \frac{ \sigma^2_{t|s} }{ \sigma_t } \mathbf{\hat{\epsilon}_{\theta}}(x_t; t)  ) \nonumber \\
& = \alpha_s \mathbf{x} +\frac{1}{\alpha_{t|s} \sigma_t} (    \sigma^2_t \mathbf{\epsilon} -    \sigma^2_{t|s} \mathbf{\hat{\epsilon}_{\theta}}(x_t; t)  ) \nonumber \\
\end{align}

Since $p(x_s|x_t) = $

\begin{align}\label{eq:qdloss}
\boldsymbol{{\mu}_{\theta}}(x_t; s, t) &=  \frac{ \alpha_{t|s} \sigma^2_{s} }{\sigma^2_{t}} \mathbf{x}_t  +  \frac{ \alpha_s \sigma^2_{t|s}}{\sigma_t^2}\mathbf {x}_0 \nonumber \\
& = \frac{ \alpha_{t|s} \sigma^2_{s} }{\sigma^2_{t}} (\alpha_t \mathbf{x}_0 + \sigma_t \mathbf{\epsilon}_t ) + \frac{ \alpha_s \sigma^2_{t|s}}{\sigma_t^2}\mathbf {x}_0 \nonumber \\
& =  \frac{ \alpha_{t} \sigma^2_{s} }{\sigma^2_{t}}  \mathbf{x}_0 + \frac{ \alpha_{t|s} \sigma^2_{s} }{\sigma_{t}}  \mathbf{\epsilon}_t +   \frac{ \alpha_s \sigma^2_{t|s}}{\sigma_t^2}\mathbf{x}_0 \nonumber \\
& = \alpha_s \mathbf{x}_0 +\frac{ \alpha_{t|s} \sigma^2_{s} }{\sigma_{t}}  \mathbf{\epsilon}_t  \nonumber \\
\end{align}

We know that the variance at time $s$, $\sigma^{2}_\theta(s, t) = \sigma^2_{t|s}\sigma^2_{s}/\sigma^2_{t}$, then we can get by sampling $p(x_s | x_t) = \mathcal{N}(x_s; \boldsymbol{{\mu}_{\theta}}(x_t; s, t), \sigma^2_\theta(s, t) \bf{I} )$
\begin{align}
\mathbf{x_s} &=  \boldsymbol{{\mu}_{\theta}}(x_t; s, t) + \sigma_\theta(s,t) \mathbf{\epsilon}_s \nonumber \\
& = \alpha_s \mathbf{x}_0 +\frac{ \alpha_{t|s} \sigma^2_{s} }{\sigma_{t}}  \mathbf{\epsilon}_t   + \sigma_\theta(s,t) \mathbf{\epsilon}_s   \nonumber \\
& = \alpha_s \mathbf{x}_0 +\frac{ \alpha_{t|s} \sigma^2_{s} }{\sigma_{t}}  \mathbf{\epsilon}_t   + \frac{\sigma_{t|s} \sigma_s }{\sigma_t} \mathbf{\epsilon}_s   \nonumber \\
\end{align}
since $\mathbf{\epsilon}_t $ and $\mathbf{\epsilon}_s $ from the same Gaussian noise, when we reduce the steps we can merge these two independent Gaussian distributions, 
the new variance can be formulated as:
\begin{align}
&(\frac{ \alpha_{t|s} \sigma^2_{s} }{\sigma_{t}})^2 + (\frac{\sigma_{t|s} \sigma_s }{\sigma_t})^2  \nonumber \\
=& \frac{ \alpha^2_{t|s} \sigma^4_{s} }{\sigma^2_{t}} + \frac{\sigma^2_{t|s} \sigma^2_s }{\sigma^2_t}  \nonumber \\
=&\frac{\sigma^2_s}{\sigma^2_t} (\alpha^2_{t|s} \sigma^2_s + \sigma^2_{t|s} ) \nonumber \\
=& {\sigma^2_s}
\end{align}

we can see that $\mathbf{x_s} \sim \alpha_s \mathbf{x}_0 + \sigma_s \epsilon$

So the most important step is to estimate accurate $\mathbf{x}$ in the inference stage. we borrow the idea from signal decomposition. The forward process of diffusion model is to add noise to the original signal until it approximate random Gaussian distribution, while the backward process is to denoise the merged the signal to recover the original data. While the data is noising, the recovered $\hat{\mathbf{x}}$, but it will be better with more denoising steps.


\bibliographystyle{unsrt}
\bibliography{dmpaper2022}

\end{document}